\documentclass{article}

 \usepackage[preprint]{neurips_2026}

\usepackage[utf8]{inputenc} 
\usepackage[T1]{fontenc}    
\usepackage{hyperref}       
\usepackage{url}            
\usepackage{booktabs}       
\usepackage{amsfonts}       
\usepackage{nicefrac}       
\usepackage{microtype}      
\usepackage{xcolor}         
\usepackage{graphicx}
\usepackage{multirow}
\usepackage{pifont}
\usepackage{booktabs}
\usepackage{xcolor}
\usepackage{wrapfig}
\usepackage{enumitem}
\usepackage{amsthm}

\usepackage{pifont}        
\usepackage{multirow}      
\usepackage{booktabs}      
\usepackage[table]{xcolor} 

\newcommand{\venue}[1]{\,{\footnotesize\textcolor{gray}{[#1]}}}

\usepackage{amsmath}

\title{SkillNav: Score-Level Skill Intervention for Zero-Shot Object Goal Navigation}

\author{
     Ruijie Sang\textsuperscript{1,$^*$} ~~Yiqun Duan\textsuperscript{2,$^*$} ~~ Pinhan Fu\textsuperscript{3,$^*$}   ~~ Ruilin Wang\textsuperscript{1} ~~ Wei Sui\textsuperscript{1,$^\dagger$} ~~Xianda Guo\textsuperscript{3,$^\ddagger$}\\
    \textsuperscript{1} D-Robotics \quad
    \textsuperscript{2} HAI Center, University of Technology Sydney\\
    \textsuperscript{3} School of Computer Science, Wuhan University\\
    sangruijie2001@163.com, ~duanyiquncc@gmail.com, ~fupinhan168@163.com\\
    ruilinking@gmail.com , ~wei.sui@d-robotics.cc, ~xianda\_guo@163.com\\
    $^*$ Equal Contributions \quad\quad $^\dagger$ Project Leader \quad\quad $^\ddagger$ Corresponding Author \\
}

\begin{document}

\maketitle

\begin{abstract}
Vision-Language Model (VLM) agents have advanced zero-shot object-goal navigation, yet single-frame reasoning leaves them without the cross-step behavioral awareness an embodied navigator requires, producing recurring failures such as dead-end stalls, in-room loops, and circuitous approaches to detected targets. Prompt-based remedies inflate token budgets across multi-submodule episodes and still struggle to encode inherently spatial signals such as angles, map cells, and viewpoint coordinates. In this paper, we propose SkillNav, an extensible behavioral skill framework for VLM-based navigation that treats the curiosity value map already maintained by modern VLM navigators as a writable substrate on which composable skills inscribe behavioral memory at zero token cost. Skills are stratified into three tiers by their level of behavioral authority, namely soft scaling for proportional reweighting, lower-bound boost for region-level guarantees, and hard override for threshold-triggered forced actions, and cooperate across tiers under a fixed composition order that establishes a predictable, declared priority among skills. This design turns capability improvement into skill registration: new behaviors plug in without retraining the VLM or disturbing existing skills, opening a path for continual refinement. A minimal prompt channel complements the score-level skills with category-level semantic hints, yielding a dual-representation design in which spatial memory lives on the map and semantic memory in short prompts. Training-free, SkillNav establishes new state-of-the-art SPL across MP3D (25.5), HM3D v0.1 (39.3), and HM3D v0.2 (43.2), improving SPL by up to 6.0 absolute over the strongest prior method, and achieves the highest Success Rate on HM3D v0.1 (69.7) and v0.2 (75.9). Code will be released upon acceptance at \url{https://github.com/XiandaGuo/SkillNav}.
\end{abstract}
\section{Introduction}
\label{sec:intro}

Object-Goal Navigation (ObjectNav) requires an embodied agent to locate an instance of a queried category in a previously unseen environment, using only egocentric observations as input~\citep{anderson2018evaluation,batra2020objectnav}. By coupling perception, semantic reasoning, and long-horizon decision making, ObjectNav serves as a canonical testbed for generalist embodied behavior and underpins downstream applications such as household service robots, search-and-rescue platforms, and assistive agents in unstructured environments~\citep{chaplot2020object,ramrakhya2022habitat}. Recent progress on this task has been driven by the repurposing of large VLMs as zero-shot, training-free reasoners for navigation~\citep{zhou2023esc,gadre2023cows,kuang2024openfmnav,liu2025toponav}. Methods such as VLFM~\citep{yokoyama2024vlfm}, SG-Nav~\citep{yin2024sgnav}, UniGoal~\citep{yin2025unigoal}, and WMNav~\citep{nie2025wmnav} cast frontier scoring, sub-goal proposal, and stop decision as VLM queries grounded in single-frame observations, while MerNav~\citep{qi2026mernav} further introduces hierarchical memory to enrich cross-step context, collectively narrowing the gap with supervised counterparts on HM3D~\citep{ramakrishnan2021habitat} and MP3D~\citep{chang2017matterport3d}.

  \begin{wrapfigure}{r}{0.5\linewidth}
  \vspace{-1.0\baselineskip}
  \centering
  \includegraphics[width=\linewidth]{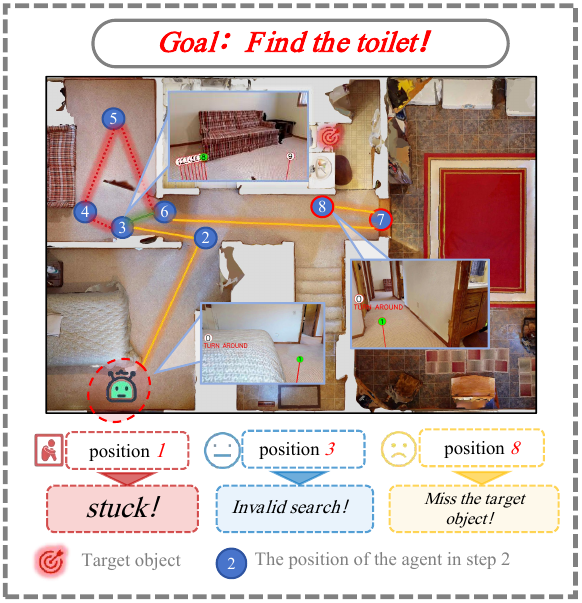}
  \caption{Three recurring failures caused by the absence of cross-step memory: getting stuck against an obstacle (\emph{position~1}), conducting invalid in-room search instead of exiting through the door (\emph{position~3}), and losing the target after it leaves the field of view (\emph{position~8}).}
  \label{fig:motivation1}
  \vspace{-0.5\baselineskip}
\end{wrapfigure}

Despite these advances, current VLM-based agents tend to fail in a recurring and stereotyped manner. As illustrated in Figure~\ref{fig:motivation1}, a robot tasked with finding a toilet may become wedged against the corner of a bed and repeatedly attempt the same failed escape directions (\emph{position~1}); it may continue to spend exploration steps deep inside a room whose appearance already rules out the target, taking a long detour rather than the short path that exits through the door (\emph{position~3}); or it may briefly catch the target in view and lose track of it after a single oversized step (\emph{position~8}). These failures share a common structural cause: the decisions are produced by a VLM that, at every step, sees only the current observation and lacks \emph{cross-step behavioral awareness} of which directions have failed, which regions are barren, and where the target was last seen.

A natural remedy, inspired by LLM agent paradigms such as ReAct~\citep{yao2023react} and Reflexion~\citep{shinn2023reflexion}, is to accumulate past observations and decisions into the textual prompt of every VLM call, so that each query reasons over the full trajectory rather than over the current frame alone~\citep{qi2026mernav,long2025instructnav,liu2025toponav}. However, this strategy carries a non-trivial cost: behavioral memory is by nature \emph{cumulative}, the prompt grows with every step, and the per-call token budget rapidly inflates across multiple reasoning sub-modules. More fundamentally, the information that one most wants to remember is intrinsically spatial, namely angles, map cells, and viewpoint coordinates rather than sentences, and a few sentences cannot faithfully encode such signals in the way that a single 2D map cell can.
 
We observe that modern VLM navigators already maintain a spatial representation, namely a curiosity value map whose cells or angular bins store scores ranking candidate motion directions~\citep{yokoyama2024vlfm,nie2025wmnav,yu2023l3mvn}. This surface offers a natural medium for behavioral memory: any spatial signal written onto it biases the next action immediately, persists across steps without re-narration, and incurs zero additional token cost. We instantiate this idea as \textbf{SkillNav}, a training-free framework inserting a \emph{Refinement Layer} between VLM-based score evaluation and action execution, monitoring the agent's cross-step behavioral state and applying structured corrections to raw direction scores. To make multiple corrections compose predictably, every intervention is abstracted into one of three \emph{modification protocols}, namely \emph{soft scaling}, \emph{lower-bound boost}, and \emph{hard override}, applied in a fixed order that makes conflicting-skill resolution explicit and predictable.
 
The contributions of this work are summarized as follows.
\begin{itemize}[leftmargin=*, align=left]
\item \textbf{Problem Identification.} We identify a structural blind spot of current VLM-based ObjectNav agents, namely the underuse of spatial information for cross-step behavioral awareness, and trace the recurrent failures of getting stuck, looping, and circuitous approach to this absence.
\item \textbf{Skill Framework.} We propose \textbf{SkillNav}, a training-free behavioral skill framework that inscribes cross-step memory on the curiosity value map at zero token cost. Skills are stratified into three tiers by authority, namely \emph{soft scaling}, \emph{lower-bound boost}, and \emph{hard override}.
\item \textbf{Composable Extensibility.} The three tiers cooperate under a fixed order under which conflicts between skills are resolved by declared authority rather than by registration accident, turning capability improvement into skill registration: new behaviors plug in by declaring their tier, opening a path for continual refinement.
\item \textbf{Results.} SkillNav consistently improves Success Rate and SPL over strong VLM-based baselines on HM3D and MP3D, establishing new SPL state of the art on MP3D (25.5), HM3D v0.1 (39.3), and HM3D v0.2 (43.2), with a 6.0-point absolute SPL improvement over the strongest prior method.
\end{itemize}
 
The remainder of this paper is as follows. Section~\ref{sec:related_work} reviews related work. Section~\ref{sec:method} presents the proposed method. Section~\ref{sec:experiments} reports experimental results and ablations. Section~\ref{sec:conclusion} concludes.

\section{Related Work}
\label{sec:related_work}

\subsection{Training-Free Object-Goal Navigation}
Existing ObjectNav methods fall into two paradigms. Supervised approaches train visual encoders coupled with RL or imitation policies~\citep{wijmans2020ddppo,ramrakhya2022habitat,maksymets2021thda,ramrakhya2023pirlnav}, or learn potential functions over semantic maps to guide exploration~\citep{chaplot2020object,ramakrishnan2022poni}, but they degrade on unseen object categories or room layouts. Zero-shot approaches remove this dependency through open-vocabulary scene understanding: image-based methods match the target against egocentric observations in a multimodal feature space~\citep{majumdar2022zson,gadre2023cows,khandelwal2022simple}, while map-based methods rank frontier candidates with LLM or VLM commonsense~\citep{yu2023l3mvn,zhou2023esc,yokoyama2024vlfm,kuang2024openfmnav,yin2024sgnav,nie2025wmnav} or operate over waypoint and topological abstractions~\citep{wu2024voronav,zhong2024topvnav}. Within the map-based zero-shot family, recent VLM-centric pipelines score frontiers~\citep{yokoyama2024vlfm,nie2025wmnav}, propose sub-goals~\citep{qi2026mernav,zhong2024topvnav}, and select actions end-to-end~\citep{goetting2025vlmnav} directly from egocentric images, avoiding the grounding gap incurred by language-mediated scene descriptions. Our framework follows this VLM-centric line but differs in what is stored on the map: rather than treating the map as a stateless scoring surface refreshed at each step, we treat it as a writable substrate that carries cross-step behavioral memory.

\subsection{Memory and State Tracking in LLM/VLM Agents}
The dominant approach to equipping LLM/VLM agents with memory is \emph{prompt-based}: past observations, actions, or sub-goals are serialized into the textual context of every model call~\citep{yao2023react,shinn2023reflexion,wang2023voyager}. Variants of this idea have been adopted in navigation for maintaining episode-level context such as exploration history and completed sub-goals~\citep{qi2026mernav,long2025instructnav,liu2025toponav}, but the prompt grows monotonically with episode length, inflating per-call token budgets and diluting attention on the current observation. An alternative line of work \emph{externalizes} memory into structured representations such as scene graphs~\citep{yin2024sgnav} or topological abstractions~\citep{wu2024voronav,zhong2024topvnav}, which the VLM then queries through prompts; however, these representations are still consumed via re-narration into language at every decision step. Our design departs from both lines: we record cross-step behavioral state directly on the curiosity value map that already drives action selection, so the memory acts on the behavior of the agent without ever being re-encoded as text.

\begin{figure}[t]
  \centering
  \includegraphics[width=0.95\linewidth]{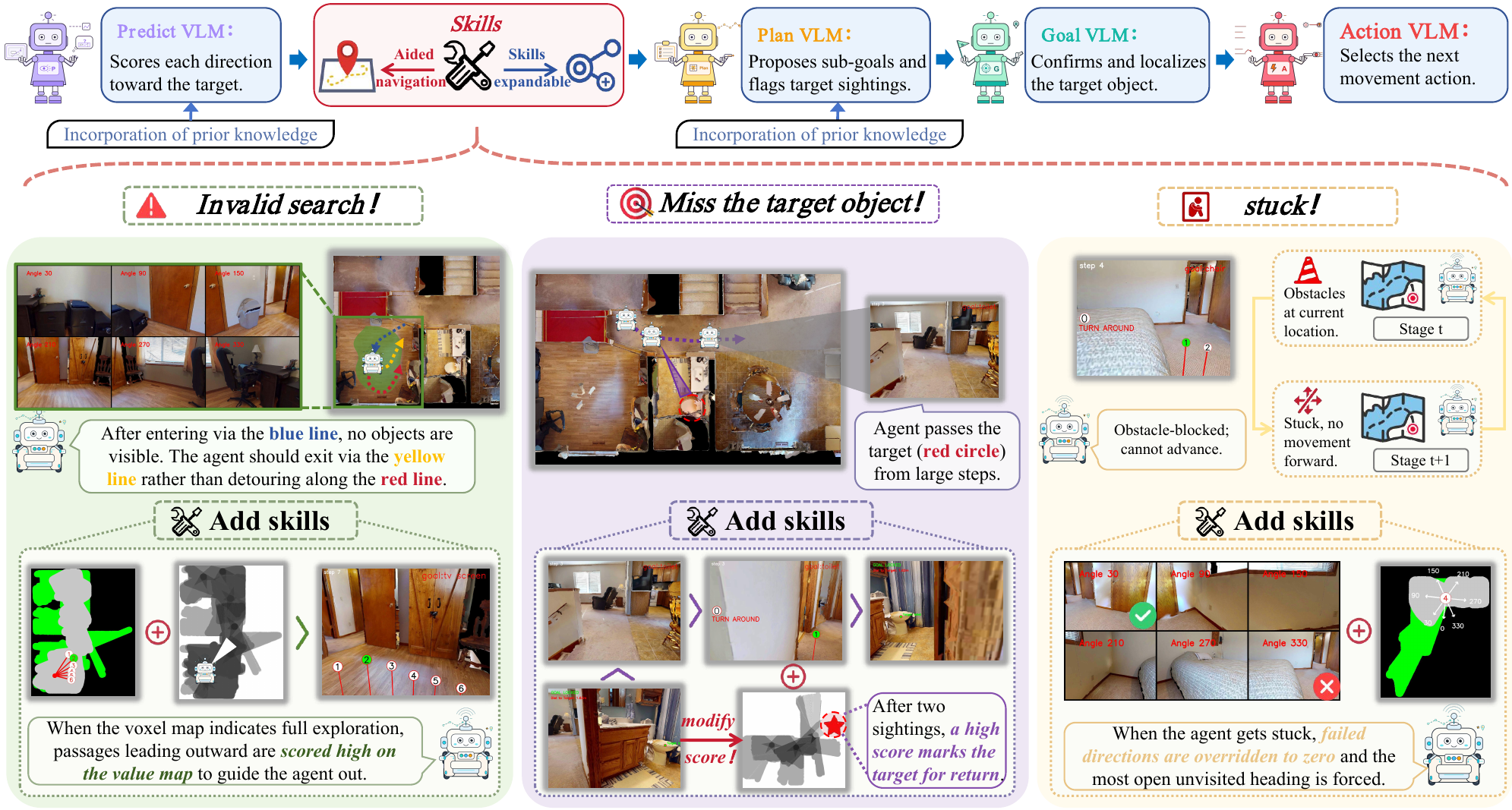}
  \caption{Overview of the SkillNav pipeline. The Predict VLM scores each heading direction, and the corrected heading is processed by the Plan, Goal, and Action VLMs. Without behavioral memory, the agent recurrently gets stuck (``stuck!''), searches an exhausted room (``Invalid search!''), or overshoots a detected target (``Miss the target object!''). The Refinement Layer resolves each by registering a skill (``Add skills'') that writes a correction onto the value map. Prior knowledge enters the Predict and Plan VLMs as bounded experience prompts.}
  \label{fig:method}
\end{figure}

\section{Method}
\label{sec:method}

This section presents SkillNav. Section~\ref{sec:method-setup} formalizes the Object-Goal Navigation task and describes the underlying VLM pipeline. Section~\ref{sec:method-refinement} introduces the \emph{Refinement Layer} and the three score-level operators it composes. Section~\ref{sec:composition} establishes the fixed composition order and its stability properties. Section~\ref{sec:method-prompt} describes the bounded prompt channel used for non-spatial signals.

\subsection{Problem Setup and Pipeline Overview}
\label{sec:method-setup}

We consider Object-Goal Navigation (ObjectNav) in unseen indoor environments. At each timestep $t$, the agent receives an egocentric RGB-D observation $o_t = (I_t, D_t)$, where $I_t \in \mathbb{R}^{H_I \times W_I \times 3}$ denotes the RGB image and $D_t \in \mathbb{R}^{H_I \times W_I}$ denotes the aligned depth map, together with its current pose. The agent then selects a discrete navigation action $a_t \in \mathcal{A}$. A target object category $g$ is specified at the start of each episode, and the episode succeeds if the agent emits a STOP action within a fixed distance threshold $d_{\text{succ}}$ of any instance of $g$. We adopt Success Rate (SR) and Success weighted by Path Length (SPL)~\citep{anderson2018evaluation} as evaluation metrics.

As illustrated in Figure~\ref{fig:method}, SkillNav augments a multi-module VLM pipeline. At the start of step $t$, the agent rotates in place to capture a panoramic split: $K$ egocentric RGB views $v_t^{(k)}$, each aligned with one of the $K$ discrete heading directions $\Theta = {\theta_1, \ldots, \theta_K}$ (we use $K=6$):
\begin{equation}
\label{eq:panorama}
    \mathcal{V}_t = \{v_t^{(1)}, \ldots, v_t^{(K)}\}, \quad v_t^{(k)} \text{ aligned with } \theta_k.
\end{equation}

\paragraph{Predict VLM.}
Given the panoramic views $\mathcal{V}_t$, the Predict VLM assigns a curiosity score $\alpha$ to each heading direction $\theta_k$, estimating how likely that direction is to lead toward an instance of $g$:
\begin{equation}
\label{eq:predict}
    \alpha(\theta_k) = f_{\text{pred}}(v_t^{(k)}, g), \alpha(\theta_k) \in [0, \alpha_{\max}].
\end{equation}
These per-direction scores are ray-cast into voxel cells and accumulated into a curiosity value map $M_t \in \mathbb{R}^{H \times W}$. Before the next heading is selected, the proposed \emph{Refinement Layer} (detailed in Section~\ref{sec:method-refinement}) applies the structured corrections of Figure~\ref{fig:method} that encode behavioral memory. The corrected map yields the next heading as $\hat\theta = \arg\max_k \tilde{\alpha}(\theta_k)$, where $\tilde{\alpha}$ denotes the refined scores.

\paragraph{Plan VLM.}
Given the chosen heading view $\hat\theta$ , the Plan VLM produces a textual sub-goal $\sigma_t$ that describes an intermediate navigation intention, together with a binary detection flag $z_t^{\text{plan}} \in \{0, 1\}$ indicating whether $g$ is visible in $\hat\theta$. The sub-goal $\sigma_t$ is forwarded to the Action VLM to condition the subsequent motion selection. If $z_t^{\text{plan}} = 1$, the Goal VLM is activated for further verification; otherwise, the pipeline proceeds directly to action selection.

\paragraph{Goal VLM.}
The Goal VLM is invoked only when $z_t^{\text{plan}} = 1$, serving as a precise verifier that either confirms the presence of $g$ and returns its pixel coordinate or rejects the sighting:
\begin{equation}
\label{eq:goal}
    \text{Goal VLM}(\hat\theta, g) = \begin{cases} (u, v) \in \mathbb{Z}^2 & \text{if } g \text{ confirmed,} \\ \emptyset & \text{otherwise.} \end{cases}
\end{equation}
A confirmed pixel $(u, v)$ is back-projected via $D_t[u, v]$ to a global 3D coordinate $\mathbf{p}^\star_t \in \mathbb{R}^3$ that anchors the location of the target for subsequent steps. If the Goal VLM rejects the sighting, no anchor is produced and the pipeline falls back to the Action VLM for continued exploration.

\paragraph{Action VLM.}
Conditioned on the sub-goal $\sigma_t$ produced by the Plan VLM and the projected forward direction, the Action VLM selects one local motion primitive $a_t \in \mathcal{A}$ to execute.

\begin{wrapfigure}{r}{0.5\linewidth}
  \vspace{-1.0\baselineskip}
  \centering
  \includegraphics[width=\linewidth]{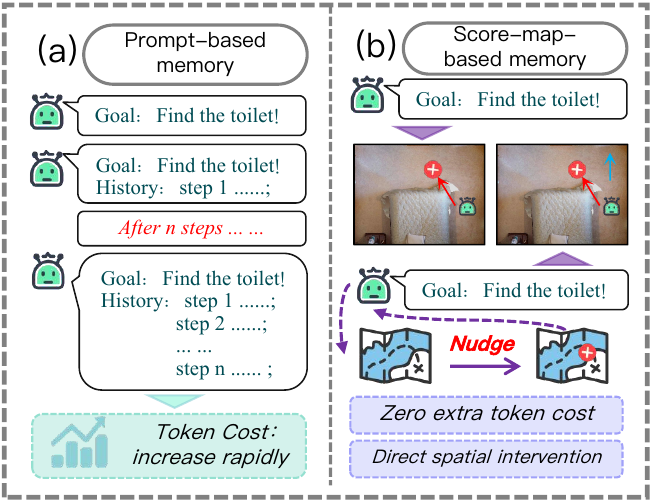}
  \caption{Two memory paradigms for VLM-based ObjectNav. (a) Prompt-based memory appends the action history to every VLM call, inflating the token cost as the episode grows. (b) SkillNav writes behavioral memory directly onto the curiosity value map and refines it through score-level skill operators, achieving direct spatial intervention at zero extra token cost.}
  \label{fig:motivation2}
  \vspace{-0.5\baselineskip}
\end{wrapfigure}

\subsection{Score-Level Skill Intervention via the Refinement Layer}
\label{sec:method-refinement}

As discussed in Section~\ref{sec:intro}, serializing the history of the agent into the textual prompt of every VLM call is a natural but costly approach to cross-step memory. As illustrated in Figure~\ref{fig:motivation2}(a), the token cost grows monotonically with episode length, inflating the per-call budget and diluting attention on the current observation. The Refinement Layer of SkillNav instead writes behavioral memory directly onto the curiosity value map, achieving direct spatial intervention at zero extra token cost (Figure~\ref{fig:motivation2}(b)). We formalize this layer as the ordered composition of three primitive operators on $\mathbb{R}^K$.

The three operator types reflect the distinct granularities at which behavioral memory needs to intervene. Some corrections are best expressed as proportional adjustments to individual direction scores, gently reweighting the existing ranking without overwriting it; this is the regime of \textbf{Soft Scaling}. Others require elevating an entire spatial region on the value map to a guaranteed minimum, regardless of what per-direction scores currently say; this is the regime of \textbf{Lower-Bound Boost}. Finally, certain failure states demand immediate, unconditional action that bypasses the scoring mechanism altogether, triggered only when monitored indicators cross predefined thresholds; this is the regime of \textbf{Hard Override}. Organizing interventions along this spectrum, from score-proportional adjustments through region-level guarantees to threshold-triggered forced actions, ensures that each behavioral correction operates at its natural level of authority and that when two corrections target the same direction, the outcome is determined by their declared tiers rather than by the order in which they happen to be registered.

\paragraph{Soft scaling.}
Given a per-direction multiplier $\gamma = (\gamma_1, \gamma_2, \ldots, \gamma_K)$ with each $\gamma_k \in (0, +\infty)$, the soft scaling operator adjusts the score of each heading direction by:
\begin{equation}
\label{eq:soft}
    S(\alpha_k) \;=\; \min\!\bigl(\gamma_k \, \alpha_k,\; \alpha_{soft}\bigr).
\end{equation}
Each score $\alpha_k$ is multiplied by $\gamma_k$ and clamped at a soft ceiling $\alpha_{\text{soft}} = 7.0$, deliberately kept below the global maximum $\alpha_{\max} = 10$ so that the top of the score range is reserved for the higher tiers, namely the region-level guarantees of Lower-Bound Boost and the forced actions of Hard Override: $\gamma_k > 1$ amplifies the score, while $\gamma_k < 1$ suppresses it without reducing it to zero. Our canonical instantiation is \emph{room exhaustion}. When the mean curiosity over visible directions falls below $\tau_{\text{exh}}$ and the voxel coverage exceeds $\tau_{\text{cov}}$, we set $\gamma_k < 1$ for already-explored directions and $\gamma_k > 1$ for directions toward unexplored frontiers or room exits, the ``Invalid search!'' remedy in Figure~\ref{fig:method}.

\paragraph{Lower-bound boost.}
This operator ensures that a confirmed target location is never forgotten due to transient score fluctuations. Whenever the Goal VLM confirms $g$ at two nearby pixels on two consecutive steps, we take $(u, v)$ to be the mean of the two confirmed pixel coordinates, back-project it via the depth value $D_t[u,v]$ to a global 3D anchor point $\mathbf{p}^\star_t$, and store it in the history buffer $h_t$. At every subsequent step, a circular region $B(\mathbf{p}^\star_t, r)$ of radius $r$ centered at $\mathbf{p}^\star_t$ is projected onto $M_t$. For each map cell $m_{i,j} \in M_t$, the boosted value $\hat{m}_{i,j}$ is defined as:
\begin{equation}
\label{eq:boost}
    \hat{m}_{i,j} = \begin{cases}\alpha_{\max} & \text{if } (i, j) \in B(\mathbf{p}^\star_t, r), \\ m_{i,j} & \text{otherwise,} \end{cases}
\end{equation}
where $m_{i,j}$ denotes the original value of cell $(i, j)$ and $\alpha_{\max}$ is the maximum allowed map value. Cells within the anchor neighborhood are thus guaranteed a value of $\alpha_{\max}$, while all other cells retain their original values unchanged. The updated map $\hat{M}_t$ subsequently influences the heading selection, ensuring that any direction whose ray passes through the anchor region retains a high score (the ``Miss the target object!'' remedy in Figure~\ref{fig:method}).

\paragraph{Hard override.}
Unlike the previous two operators, hard override completely disregards the original scores and forces the agent to execute a specific action when certain behavioral indicators reach critical thresholds. For each heading direction $\theta_k$, the overridden score is:
\begin{equation}
\label{eq:override}
    \hat{\alpha}_k = \begin{cases} c_k & k \in \mathcal{O}, \\ \alpha_k & k \notin \mathcal{O}, \end{cases}
\end{equation}
where $\mathcal{O} \subseteq {1, \ldots, K}$ denotes the set of overridden directions and $c_k$ the replacement value. Our canonical instantiation is \emph{anti-stuck recovery}. The agent monitors its $L_2$ displacement over a sliding window of $W{=}2$ steps, and activates an escape sub-routine when it falls below $\tau_{\text{move}}$. The sub-routine identifies the most open heading $\theta^\dagger$ disjoint from previously failed directions $\mathcal{F}_t$ and constructs:
\begin{equation}
\label{eq:override_set}
    \mathcal{O} = \{k^\dagger\} \cup \mathcal{F}_t, \quad c_k = \begin{cases} c_{\text{esc}} & k = k^\dagger, \\ 0 & k \in \mathcal{F}_t. \end{cases}
\end{equation}
If $\theta^\dagger$ also fails to produce movement, $k^\dagger$ is appended to $\mathcal{F}_t$ and the sub-routine selects the next most open direction, progressively eliminating dead ends until a viable exit is found. In addition, an ABAB oscillation detector checks whether the agent alternates between two positions over the last four steps; when triggered, the directions toward the repeated waypoint are added to $\mathcal{F}_t$ and overridden to $0$, breaking the loop (the ``stuck!'' remedy in Figure~\ref{fig:method}).

\subsection{Composition and Stability Guarantee}
\label{sec:composition}

The three operators require principled ordering so that, when skills act on the same direction, the outcome follows from their declared tiers. The Refinement Layer applies them in this fixed sequence:
\begin{equation}
\label{eq:compose}
    \mathcal{R}(\alpha, M_t, h_t) \;=\; \text{Hard Override} \circ \text{Lower-Bound Boost} \circ \text{Soft Scaling}\,(\alpha),
\end{equation}
that is, soft scaling is applied first, followed by lower-bound boost, and finally hard override. After composition writes final scores to $M_t$, the heading is selected as $\hat\theta = \arg\max_k \hat{\alpha}_k$. The Plan, Goal, and Action VLMs remain unmodified, making SkillNav compatible with existing pipelines.

\paragraph{Stability Properties.}
This ordering is the unique one, up to permutations among operators with disjoint active supports, that simultaneously satisfies the following three properties.

(P1) Floor preservation: let $\beta_k \geq 0$ denote the lower-bound floor assigned to direction $k$ by the boost operator. Then for every $k$ with $\beta_k > 0$ and $k \notin \mathcal{O}$, the final score satisfies $\hat{\alpha}_k \geq \beta_k$.

(P2) Override dominance: for every direction $k \in \mathcal{O}$, the final score satisfies $\hat{\alpha}_k = c_k$.

(P3) Scaling neutrality: if $\gamma_k = 1$, $\beta_k = 0$, and $k \notin \mathcal{O}$, then $\hat{\alpha}_k = \alpha_k$.

\paragraph{Intuition.}
Applying soft scaling after lower-bound boost violates (P1), because a multiplier $\gamma_k < 1$ can demote a boosted direction below its floor. Applying lower-bound boost after hard override violates (P2), because a non-zero floor can re-elevate a direction just overridden to $0$.

\paragraph{Practical Consequence.}
New behavioral skills can be integrated into SkillNav by specifying only their operator family (scaling, boost, or override), without reasoning about interactions with existing skills. The fixed composition order guarantees conflict resolution is fully determined by (P1)--(P3), so a newly registered skill can never alter the semantics of an existing one unexpectedly.

\subsection{Experience Prompt Injection}
\label{sec:method-prompt}

The Refinement Layer corrects behavioral failures via spatial encodings, but a residual class of errors arises from perceptual ambiguity: VLMs may confuse a stool with a chair, a couch with a bed, or accept a mirror reflection as a valid detection. As shown by the prior-knowledge channels in Figure~\ref{fig:method}, SkillNav addresses this through a bounded experience prompt channel. At episode start, at most $L_{\max}$ category-specific hints for $g$ are loaded from an offline library and appended to the system prompts of the Predict and Plan VLMs, encoding cues such as \emph{a chair must have a backrest} and \emph{objects visible only in mirrors are not valid detections}. These hints remain fixed throughout the episode, incurring a token cost of $O(1)$ in episode length.

No position, bearing, or per-step decision ever enters this channel. All spatial behavioral memory is handled exclusively by the Refinement Layer through $M_t$ and $h_t$, while the prompt channel carries only category-level disambiguation cues. This strict separation keeps the two representations complementary: the score map records \emph{where} the agent has been, while the experience prompt clarifies \emph{what} it is looking for.
\section{Experiments}
\label{sec:experiments}

This section evaluates SkillNav. Section~\ref{sec:exp-setup} describes the experimental setup, covering datasets, metrics, and implementation details. Section~\ref{sec:exp-main} compares SkillNav against zero-shot and supervised baselines on HM3D v0.1, HM3D v0.2, and MP3D. Section~\ref{sec:exp-ablation} ablates the prompt channel and the Refinement Layer, showing that the two mechanisms target distinct failures. Section~\ref{sec:exp-percat} provides a per-category breakdown across two VLM backbones and discusses remaining bottlenecks.

\subsection{Experimental Setup}
\label{sec:exp-setup}

We evaluate on three standard ObjectNav benchmarks: \textbf{HM3D v0.1}, \textbf{HM3D v0.2}~\citep{ramakrishnan2021habitat,yadav2023habitat} (6 target categories), and \textbf{MP3D}~\citep{chang2017matterport3d} (21 target categories). All experiments run inside the Habitat simulator with a fixed budget of 40 steps, following the evaluation setting of WMNav~\citep{nie2025wmnav}, and a success threshold of $d_{\text{succ}} = 1.0$\,m. We report Success Rate (SR) and Success weighted by Path Length (SPL) following the standard protocol~\citep{anderson2018evaluation}. Our main results use \textbf{Gemini-3-Flash-Preview} as the VLM backbone; ablations are additionally conducted with \textbf{Qwen2.5-VL-7B-Instruct} to verify backbone independence. SkillNav is fully training-free. 

\begin{table*}[!t]
\centering
\caption{Comparison with state-of-the-art methods on ObjectNav. TF: training-free, ZS: zero-shot. Best results are in \textbf{bold}, second best are \underline{underlined}.}
\label{tab:main_results}
\vspace{6pt}
\small
\setlength{\tabcolsep}{8pt}
\begin{tabular}{l|| cc| cc |cc |cc}
\toprule
\rowcolor[gray]{0.92}
& & & \multicolumn{2}{c|}{\textbf{MP3D}} & \multicolumn{2}{c|}{\textbf{HM3D v0.1}} & \multicolumn{2}{c}{\textbf{HM3D v0.2}} \\
\rowcolor[gray]{0.92}
\multirow{-2}{*}{\textbf{Method}} & \multirow{-2}{*}{\textbf{TF}} & \multirow{-2}{*}{\textbf{ZS}}
& SR & SPL & SR & SPL & SR & SPL \\
\midrule
ZSON\venue{NeurIPS} \cite{majumdar2022zson}            & \ding{55} & \ding{51} & 15.3 & 4.8  & 25.5 & 12.6 & --   & --   \\
PixNav\venue{ICRA} \cite{cai2024bridging}                & \ding{55} & \ding{51} & --   & --   & 37.9 & 20.5 & --   & --   \\
PSL\venue{ECCV} \cite{sun2024psl}                      & \ding{55} & \ding{51} & 18.9 & 6.4  & 42.4 & 19.2 & --   & --   \\
SGM\venue{CVPR} \cite{zhang2024sgm}                    & \ding{55} & \ding{51} & 37.7 & 14.7 & 60.2 & 30.8 & --   & --   \\
VLFM\venue{ICRA} \cite{yokoyama2024vlfm}               & \ding{55} & \ding{51} & 36.4 & 17.5 & 52.5 & 30.4 & 62.6 & 31.0 \\
\midrule
CoW\venue{CVPR} \cite{gadre2023cows}                    & \ding{51} & \ding{51} & 9.2  & 4.9  & --   & --   & --   & --   \\
ESC\venue{ICML} \cite{zhou2023esc}                     & \ding{51} & \ding{51} & 28.7 & 14.2 & 39.2 & 22.3 & --   & --   \\
L3MVN\venue{IROS} \cite{yu2023l3mvn}                   & \ding{51} & \ding{51} & --   & --   & 50.4 & 23.1 & 36.3 & 15.7 \\
VoroNav\venue{ICML} \cite{wu2024voronav}               & \ding{51} & \ding{51} & --   & --   & 42.0 & 26.0 & --   & --   \\
TopV-Nav\venue{arXiv} \cite{zhong2024topvnav}          & \ding{51} & \ding{51} & 35.2 & 16.4 & 53.0 & 29.8 & --   & --   \\
OpenFMNav\venue{ICLR} \cite{kuang2024openfmnav}        & \ding{51} & \ding{51} & --   & --   & 54.9 & 24.4 & --   & --   \\
SG-Nav\venue{NeurIPS} \cite{yin2024sgnav}              & \ding{51} & \ding{51} & 40.2 & 16.0 & 54.0 & 24.9 & 49.6 & 25.5 \\
VLMNav\venue{NeuS} \cite{goetting2025vlmnav}           & \ding{51} & \ding{51} & --   & --   & 50.4 & 21.0 & --   & --   \\
UniGoal\venue{CVPR} \cite{yin2025unigoal}              & \ding{51} & \ding{51} & 41.0 & 16.4 & 54.5 & 25.1 & --   & --   \\
InstructNav\venue{CoRL} \cite{long2025instructnav}     & \ding{51} & \ding{51} & --   & --   & 58.0 & 20.9 & --   & --   \\
MFNP\venue{arXiv} \cite{zhang2025mfnp}                 & \ding{51} & \ding{51} & 41.1 & 15.4 & 58.3 & 26.7 & --   & --   \\
TopoNav\venue{arXiv} \cite{liu2025toponav}             & \ding{51} & \ding{51} & 45.5 & 16.8 & 60.1 & 34.6 & --   & --   \\
WMNav\venue{IROS} \cite{nie2025wmnav}                  & \ding{51} & \ding{51} & 45.4 & 17.2 & 58.1 & 31.2 & 72.2 & 33.3 \\
MerNav\venue{arXiv} \cite{qi2026mernav}                & \ding{51} & \ding{51} & \textbf{50.8} & \underline{19.5} & \underline{68.0} & \underline{36.9} & \underline{74.8} & \underline{37.6} \\
\midrule
\rowcolor{cyan!10}
\textbf{SkillNav (Ours)} & \ding{51} & \ding{51} & \underline{48.3} & \textbf{25.5} & \textbf{69.7} & \textbf{39.3} & \textbf{75.9} & \textbf{43.2} \\
\bottomrule
\end{tabular}
\end{table*}

\begin{table}[!t]
\centering
\caption{Ablation study results on HM3D v0.2. Best results are in \textbf{bold}.}
\label{tab:ablation}
\vspace{6pt}
\small
\setlength{\tabcolsep}{6pt}
\begin{tabular}{c l || cc |cc |cc}
\toprule
\rowcolor[gray]{0.92}
& & & & \multicolumn{2}{c}{\textbf{Navigation}} & \multicolumn{2}{c}{\textbf{Efficiency}} \\
\rowcolor[gray]{0.92}
\multirow{-2}{*}{\textbf{\#}} & \multirow{-2}{*}{\textbf{Model}} & \multirow{-2}{*}{\textbf{Memory}} & \multirow{-2}{*}{\textbf{Skill}} & SR$\uparrow$ & SPL$\uparrow$ & FP$\downarrow$ & MS$\downarrow$ \\
\midrule
a & Qwen2.5-VL-7B-Instruct & \ding{55} & \ding{55} & 69.9 & 28.3 & 10.1 & 19.8 \\
b & Qwen2.5-VL-7B-Instruct & \ding{51} & \ding{55} & 70.5 & 28.1 & \textbf{8.9} & 20.4 \\
c & Qwen2.5-VL-7B-Instruct & \ding{51} & \ding{51} & 71.6 & 31.9 & 9.7 & 18.7 \\
\midrule
\rowcolor{cyan!10}
d & Gemini-3-Flash-Preview & \ding{51} & \ding{51} & \textbf{75.9} & \textbf{43.2} & 10.7 & \textbf{13.4} \\
\bottomrule
\end{tabular}
\end{table}

\subsection{Main Results}
\label{sec:exp-main}

\paragraph{State-of-the-art SPL across all three benchmarks.}
As shown in Table~\ref{tab:main_results}, SkillNav establishes new SPL state of the art on every benchmark: 25.5 on MP3D, 39.3 on HM3D v0.1, and 43.2 on HM3D v0.2, surpassing the strongest prior method (MerNav) by $+6.0$, $+2.4$, and $+5.6$ absolute, respectively. We note that a subset of episodes in HM3D v0.1 contain annotation or navigation-graph issues that cause inconsistent success judgments; the results reported in Table~\ref{tab:main_results} are computed after filtering these problematic instances. The gap is largest on MP3D, the benchmark with the most diverse category set and the longest geodesic distances, where path efficiency is hardest to maintain. This pattern is consistent with our central claim: writing cross-step behavioral memory directly onto the value map prevents the detours and re-visits that uniformly degrade SPL across baselines.

\paragraph{Highest SR on HM3D, competitive on MP3D.}
SkillNav achieves the best SR among zero-shot training-free methods on HM3D v0.1 (69.7, $+1.7$ over MerNav) and HM3D v0.2 (75.9, $+1.1$ over MerNav). On MP3D, SkillNav reaches 48.3, slightly below the 50.8 of MerNav but accompanied by a 6.0-point SPL lead. This trade-off reflects the design: methods that maximize SR on MP3D tend to extend exploration aggressively, inflating path length, whereas the behavioral memory of SkillNav short-circuits unnecessary exploration once a target is anchored, sacrificing a small fraction of late-episode successes for a substantial gain in path efficiency. Beyond the training-free group, SkillNav also surpasses all supervised zero-shot baselines on both metrics across all three benchmarks.

\subsection{Ablation Study}
\label{sec:exp-ablation}

We ablate the two memory mechanisms in SkillNav, the bounded prompt channel (\textbf{Memory}) and the Refinement Layer (\textbf{Skill}), on HM3D v0.2 (Table~\ref{tab:ablation}). Row (a) is the baseline pipeline with neither mechanism; rows (b) and (c) add Memory and Skill cumulatively; row (d) replaces the Qwen2.5-VL backbone with Gemini-3-Flash.

\paragraph{Each module contributes positively, and the gains are additive.}
Adding Memory alone (a $\to$ b) raises SR from 69.9 to 70.5 ($+0.6$). Adding Skill on top (b $\to$ c) further raises SR to 71.6 and SPL from 28.1 to 31.9 ($+3.8$ absolute). The full configuration improves on the baseline by $+1.7$ SR and $+3.6$ SPL, confirming that the two mechanisms are complementary.

\paragraph{Memory targets FP, the Refinement Layer targets MS.}
The failure-mode breakdown reveals a sharper picture. Adding Memory alone cuts the false-positive rate from 10.1\% to 8.9\% ($-1.2$) while slightly increasing the max-steps rate ($19.8 \to 20.4$). Adding Skill on top reverses the pattern: FP drifts back up slightly ($8.9 \to 9.7$), but MS drops sharply from 20.4 to 18.7 ($-1.7$). The two mechanisms therefore resolve different failure modes: the prompt channel injects category-level commonsense, such as \emph{a chair has a backrest, a stool does not}, which helps the agent reject look-alike confusions and avoid premature stops, while the score-level corrections of the Refinement Layer eliminate the stuck-loop-detour patterns of Section~\ref{sec:intro} that otherwise cause episodes to time out. This empirical decoupling directly supports the dual-representation principle motivated in Section~\ref{sec:method-prompt}: semantic memory is the right tool for semantic confusions, spatial memory is the right tool for spatial inefficiencies, and forcing one to do the job of the other, as in prior prompt-only or map-only designs, leaves one of the two failure modes unaddressed.

\paragraph{Stronger backbones lift the ceiling without changing the structure.}
Replacing Qwen2.5-VL with Gemini-3-Flash (c $\to$ d) raises SR from 71.6 to 75.9 and SPL from 31.9 to 43.2. The MS rate drops further ($18.7 \to 13.4$), while FP rises modestly ($9.7 \to 10.7$). The same SkillNav framework therefore continues to extract gains from a stronger VLM, and the residual failure mass shifts further toward semantic confusion (FP) rather than navigational inefficiency (MS), consistent with our reading that SkillNav has saturated the structural side of the problem.

\subsection{Per-Category Analysis}
\label{sec:exp-percat}

To localize the remaining bottlenecks, Table~\ref{tab:per_category} reports SR and SPL per target category for the two backbones used in the ablation.

\begin{table*}[!t]
\centering
\caption{Per-category ObjectNav results on HM3D v0.2 with different backbone models.}
\label{tab:per_category}
\vspace{6pt}
\small
\setlength{\tabcolsep}{8pt}
\begin{tabular}{l ccc | l ccc}
\toprule
\rowcolor[gray]{0.92}
\multicolumn{4}{c|}{\textbf{Qwen2.5-VL-7B-Instruct}} & \multicolumn{4}{c}{\textbf{Gemini-3-Flash-Preview}} \\
\rowcolor[gray]{0.92}
\textbf{Object} & \textbf{SR} & \textbf{SPL} & \textbf{Succ / Total} & \textbf{Object} & \textbf{SR} & \textbf{SPL} & \textbf{Succ / Total} \\
\midrule
Bed        & 75.2 & 35.9 & 124 / 165 & Bed       & 85.5 & 47.6 & 141 / 165 \\
Sofa       & 81.8 & 42.7 & 153 / 187 & Sofa      & 81.8 & 49.3 & 153 / 187 \\
Chair      & 83.1 & 35.2 & 162 / 195 & Chair     & 79.0 & 43.2 & 154 / 195 \\
Toilet     & 75.3 & 35.0 & 125 / 166 & Toilet    & 75.3 & 46.4 & 125 / 166 \\
TV Screen  & 57.0 & 24.4 & ~77 / 135 & TV Screen & 73.3 & 43.5 & ~99 / 135 \\
Plant      & 49.3 & 13.0 & ~75 / 152 & Plant     & 57.2 & 27.4 & ~87 / 152 \\
\midrule
\rowcolor{cyan!10}
\textbf{Overall} & \textbf{71.6} & \textbf{31.9} & \textbf{716 / 1000} & \textbf{Overall} & \textbf{75.9} & \textbf{43.2} & \textbf{759 / 1000} \\
\bottomrule
\end{tabular}
\vspace{-10pt}
\end{table*}

\paragraph{Visually Salient Categories Already Saturate.}
The two backbones differ little in SR on visually unambiguous categories: Sofa (81.8 vs.\ 81.8) and Toilet (75.3 vs.\ 75.3) are essentially flat, while Chair is a mild exception in the opposite direction, with the stronger backbone slightly lowering SR (83.1 to 79.0, $-4.1$). On this group, Gemini's gain appears in SPL rather than SR ($+8.0$ on Chair, $+6.6$ on Sofa, $+11.4$ on Toilet): on episodes that succeed, a stronger VLM reaches the target along shorter trajectories, even though it unlocks no additional successes and, on Chair, converts a few prior successes into failures.

\paragraph{Visually Ambiguous Categories Carry the Gap.}
The pattern reverses on categories prone to perceptual confusion. TV Screen exhibits the largest SR gain ($+16.3$, from 57.0 to 73.3), followed by Bed ($+10.3$) and Plant ($+7.9$), with equally pronounced SPL improvements ($+19.1$ on TV Screen, $+14.4$ on Plant). These are targets most susceptible to mistaken identity---a powered-off screen, a bed conflated with a sofa, a small plant lost on a cluttered shelf---confirming that the bottleneck lies at the VLM's perceptual side rather than at the navigation policy.

\paragraph{Implications for Future Work.}
The chair/sofa/toilet plateau shows SkillNav has substantially closed the cross-step behavioral awareness gap on perceptually easy targets, while the dominant residual failure source is visual grounding on ambiguous ones, a regime most amenable to improvement by scaling the underlying VLM. Two directions emerge: a goal verification module re-checking candidates across multiple viewpoints before committing the stop action, and a hierarchical experience prompt sharing commonsense spatial priors across semantically related categories, relaxing the per-category prompt budget without violating the constant-token-cost constraint of Section~\ref{sec:method-prompt}.

\section{Conclusion}
\label{sec:conclusion}

This paper identifies underuse of spatial information as a structural blind spot of VLM-based ObjectNav agents and proposes SkillNav, an extensible behavioral skill framework inscribing memory directly on the curiosity value map. Skills are stratified into three tiers by behavioral authority: soft scaling, lower-bound boost, and hard override, and cooperate under a fixed composition order resolving inter-skill conflicts by declared authority, complemented by a bounded experience prompt channel for category-level disambiguation. Together, they target orthogonal failures and establish new SPL state of the art on all three benchmarks and the highest SR on HM3D among training-free zero-shot methods, while reframing capability improvement as skill registration for continual refinement.


\newpage
\bibliography{main}
\bibliographystyle{abbrvnat}


\end{document}